# A Tableaux Calculus for Ambiguous Quantification[*]

Christof Monz and Maarten de Rijke

ILLC, University of Amsterdam, Plantage Muidergracht 24, 1018 TV Amsterdam, The Netherlands. E-mail: {christof, mdr}@wins.uva.nl

**Abstract.** Coping with ambiguity has recently received a lot of attention in natural language processing. Most work focuses on the semantic representation of ambiguous expressions. In this paper we complement this work in two ways. First, we provide an entailment relation for a language with ambiguous expressions. Second, we give a sound and complete tableaux calculus for reasoning with statements involving ambiguous quantification. The calculus interleaves partial disambiguation steps with steps in a traditional deductive process, so as to minimize and postpone branching in the proof process, and thereby increases its efficiency.

## 1 Introduction

Natural language expressions can be highly ambiguous, and this ambiguity may have various faces. Well-known phenomena include lexical and syntactic ambiguities. In this paper we focus on representing and reasoning with a different source of ambiguity, namely quantificational ambiguity, as exemplified in (1).

(1) a. Every man loves a woman.
    b. Every boy doesn't see a movie.

The different readings of (1.a) correspond to the two logical representations in

(2) a. $\forall x\,(man(x) \to \exists y\,(woman(y) \land love(x,y)))$.
    b. $\exists y\,(woman(y) \land \forall x\,(man(x) \to love(x,y)))$.

We refer the reader to [KM93,DP96] for extensive discussions of these and other examples of quantificational ambiguity. All we want to observe here is this. Examples like (1.a) have a preferred reading namely the wide-scope reading represented by (2.a)). Additional linguistic or non-linguistic information, or the context, may overrule this preference. For instance, if (1.a) is followed by (3), then the second reading (2.b) is preferred. But if (1.a) occurs in isolation, then the first reading (2.a) is preferred.

(3) But she is already married.

Clearly, if we want to process a discourse from left to right and take the context of an expression into account, our semantic representation for (1.a) must initially allow for both possibilities. And, similarly, any reasoning system for ambiguous expressions needs to be able to integrate information that helps the disambiguation process within the deductive process.

[*] The research in this paper was supported by the Spinoza project 'Logic in Action' at the University of Amsterdam.

Although the problem of ambiguity and underspecification has recently enjoyed a considerable increase in attention from computational linguists, computer scientists and logicians (see, for instance, [DP96]), the focus has mostly been on semantic aspects, and deductive reasoning with ambiguous sentences is still in its infancy.

The aim of this paper is to present a tableaux calculus for reasoning with expressions involving ambiguous quantification. An important feature of our calculus is that it integrates two processes: disambiguation and deductive reasoning. The calculus operates on semantic representations of natural language expressions. These representations contain both ambiguous and unambiguous subparts, and an important feature of our representations is that they represent all possible disambiguations of an ambiguous statement in such a way that unambiguous subparts are shared as much as possible. As we will explain below, compact representations of this kind will allow us to keep ambiguities 'localized' — a feature which has important advantages from the point of view of efficiency.

In setting up a deductive system for ambiguous quantification we have had two principal desiderata. First, although this is not the topic of the present paper, we aim to implement the calculus as part of a computational semantics work bench; this essentially limits our options to resolution and tableaux based calculi. Second, to incorporate information arising from the disambiguation process within a proof system, the proofs themselves need to be incremental in the sense that at any stage we have a 'partial' proof that can easily be extended to cope with novel information. We believe that a tableaux style calculus has clear advantages over resolution based systems in this respect.

The paper is organized as follows. A considerable amount of work goes into setting up semantic representations and a mechanism for for recording ambiguities and disambiguations in such a way that it interfaces rather smoothly with traditional deductive proof steps. This work takes up Sections 2 and 3. Then, in Section 4 we present two tableaux calculi, one which deals with fully disambiguated representations of ambiguous natural language expressions, and a more interesting one in which traditional tableaux style deduction is interleaved with partial disambiguation. Section 5 contains a detailed example, and Section 6 provides conclusions and suggestions for further work.

## 2   Representing Ambiguity

Lexical ambiguities can be represented pretty straightforwardly by putting the different readings into a disjunction. (Cf. [Dee96,KR96] for further elaboration.) It is also possible to express quantificational ambiguities by a disjunction, but quite often this involves much more structure than in the case of lexical ambiguities, because quantificational ambiguities are not tied to a particular atomic expression. For instance, the only way to represent the ambiguity of (1.a) in a disjunctive manner is (4).

(4)   $\forall x\,(man(x) \to \exists y\,(woman(y) \wedge love(x,y)))$
      $\vee \exists y\,(woman(y) \wedge \forall x\,(man(x) \to love(x,y)))$

Obviously, there seems to be some redundancy, because some subparts appear twice. If we put indices at the corresponding subparts, as in (5) below, we see that these subparts are not proper expressions of first-order logic, except subpart $k$.

(5) $\underline{\forall x\,(man(x) \to\,}_i \underline{\exists y\,(woman(y) \land\,}_j \underline{love(x,y)\,}_k))$
$\lor \underline{\exists y\,(woman(y) \land\,}_j \underline{\forall x\,(man(x) \to\,}_i \underline{love(x,y)\,}_k))$

The difference between the readings lies not in the material used, both readings are built from the parts $i$, $j$ and $k$, but in the order these are put together.

A reasonable way to represent improper expressions like $i$ and $k$ is to abstract over those parts that are missing in order to yield a proper expression of first-order logic. [Bos95] calls these missing parts *holes*. Roughly speaking, they are variables over occurrences of first-order formulas. To distinguish the occurrence of an expression from its logical content, it is necessary to supplement first-order formulas with labels. Holes may be subject to constraints; for instance, the semantic representations of verbs have to be in the scope of its arguments, because otherwise it may happen that the resulting disambiguations contain free variables. So we do not want to permit disambiguations like $\forall x\,(man(x) \to love(x,y) \land \exists y\,(woman(y)))$. These constraints are expressed by a partial order on the labels.

**Definition 1 (Underspecified Representation).** For $i \in \mathbb{N}$, let $h_i$ a new atomic symbol, called a *hole*. A formula $\varphi$ is an *h-formula*, or a formula possibly containing holes, if it is built up from holes and atomic formulas from first-order logic using the familiar boolean connectives and quantifiers.

Next, we specify the format of an *underspecified representation UR* of a natural language expression. An underspecified representation is a quadruple $\langle LHF, L, H, C \rangle$ consisting of

1. A set of labeled h-formulas *LHF*.
2. The set of labels $L$ occurring in *LHF*.
3. The set of holes $H$ occurring in *LHF*.
4. A set of order-constraints $C$ of the form $k \leq k'$, meaning that $k$ has to be a subexpression of $k'$, where $k, k' \in L \cup H$ and $C$ is closed under reflexivity, antisymmetry and transitivity.

An obvious question at this point is, how does one associate a *UR* with a given natural language expression? We will not address this issue here, but we will assume that there exists some mechanism for arriving at *UR*'s, see for example [Kön94]. For notational convenience we write $UR(S)$ for the underspecified representation, associated with a sentence $S$. By way of example, we reconsider (4) and obtain the following underspecified representation:

(6) $\langle \{l_0 : h_0, l_1 : \forall x\,(man(x) \to h_1), l_2 : \exists y\,(woman(y) \land h_2), l_3 : love(x,y)\},$
  $\{l_0, l_1, l_2, l_3\},$
  $\{h_0, h_1, h_2, h_3\}\rangle,$
  $closure(\{l_1 \leq h_0, l_2 \leq h_0, l_3 \leq h_1, l_3 \leq h_2\})$

There are two possible sets of instantiations, $\iota_1$ and $\iota_2$, of the holes $h_0, h_1, h_2, h_3$ in (6) which obey the constraints in (6): $\iota_1 = \{h_0 := l_1, h_1 := l_2, h_2 := l_3\}$ and $\iota_2 = \{h_0 := l_2, h_2 := l_1, h_1 := l_3\}$.
It is also possible to view *UR*'s as upper semi-lattices, as it is done in [Rey93]:

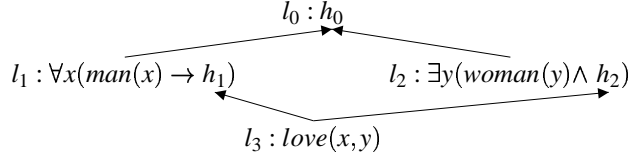

For each instantiation of the holes there is a corresponding substitution $\sigma(\iota)$ which is like $\iota$ but $h := \varphi \in \sigma(\iota)$ iff there is a $l$, such that $l : \varphi \in LHF$ and $h := l \in \iota$.

The next step is to define an extension of the language of first-order logic, $\mathcal{L}$, in which both standard (unambiguous) expressions occur side by side with the above underspecified representations. The resulting language of the language of underspecified logic, or $\mathcal{L}^u$ for short, is the language in which we will perform deduction.

**Definition 2 (Underspecified Logic).** A formula $\varphi$ is a formula of our *underspecified logic* $\mathcal{L}^u$, or a *u-formula*, that is, a formula possibly containing underspecified representations, if it is built up from underspecified representations and the usual atomic formulas from standard first-order logic using the familiar boolean connectives and quantifiers.

*Example 3.* As an example of a more complex u-formula consider the semantic representation of *if every boy didn't sleep and John is a boy, then John didn't sleep*.

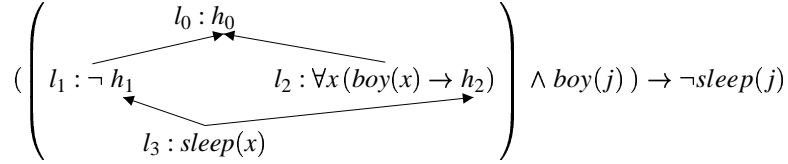

**Definition 4 (Total Disambiguations).** To define the total disambiguation $\delta(\varphi)$ of a u-formula $\varphi$, we need the following notion of a *join*.

Given an underspecified representation $\langle LHF, L, H, C \rangle$ and $k, k', k'' \in L \cup H$ and $k'' \leq k, k' \in C$ then $k''$ is the *join* of $k$ and $k'$, $k \sqcup k' = k''$, only if there is no $k''' \in L \cup H$ and $k''' \leq k, k' \in C$ and $k''' > k'' \in C$.

Then, by $\delta(\varphi)$ we denote the set of total disambiguations of the u-formula $\varphi$, where for all $d \in \delta(\varphi)$, $d \in \mathcal{L}$. For complex u-formulas $\delta$ is defined recursively:

1. $\delta(\langle LHF, L, H, C \rangle) =$ the set of $LHF\sigma(\iota)$ such that
    (i) $\iota$ is an instantiation and $\sigma(\iota)$ is the corresponding substitution
    (ii) $H\iota = L$
    (iii) for all $l, l' \in L$, if $l \sqcup l'$ is defined, then $l \leq l' \in closure(C\iota)$ or $l' \leq l \in closure(C\iota)$
2. $\delta(\neg \varphi) = \{ \neg d \mid d \in \delta(\varphi) \}$
3. $\delta(\varphi \circ \psi) = \{ d \circ d' \mid d \in \delta(\varphi), d' \in \delta(\psi) \}$, where $\circ \in \{\wedge, \vee, \rightarrow\}$
4. $\delta(Qx\varphi) = \{ Qxd \mid d \in \delta(\varphi) \}$, where and $Q \in \{\forall, \exists\}$.

If $l \leq l' \notin C$ and $l' \leq l \notin C$, then it does not have to be case that there is a scope ambiguity between quantifiers belonging to $l$ and $l'$. For instance, if $l$ and $l'$ belong to different conjuncts, they are not ordered to each other. The restriction that $l \sqcup l'$ has to be defined excludes this.

*Example 5.* To illustrate the purpose of this restriction see the underspecified representation for *every man who doesn't have a car rides a bike*

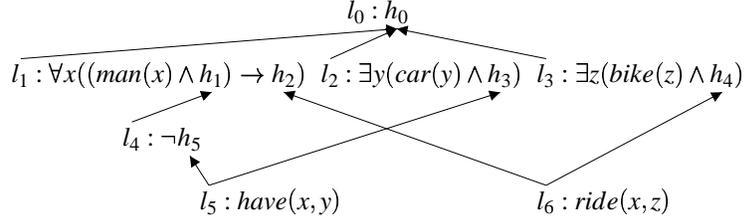

Although $l_3$ and $l_4$ are not related to each other, it cannot happen that $l_3$ is in the scope of $l_4$, because the negation must be a subformula of the antecedent of $l_1$, whereas $l_3$ might have scope over $l_1$ as a whole or might be in the scope of the succedent of $l_1$. More generally, this is due to the fact that $l_3$ and $l_4$ do not have to share a subformula, i.e., $l_3 \sqcup l_4$ is not defined.

## 3 Semantics of Underspecified Formulas

In the previous section we introduced a formalism that allows for a compact semantic representation of ambiguous expressions. Now we want to see what the validity conditions of these underspecified representations are, and how they interact with the classical logical connectives.

If an ambiguous sentence $S$ with $\delta(UR(S)) = \{d_1, d_2\}$ is uttered, and we want to check, whether $S$ is valid, we simply have to see whether all of its disambiguations are valid. That is, it must be the case that $\models d_1$ and $\models d_2$. If, on the other hand, an ambiguous sentence $S$ with $\delta(UR(S)) = \{d_1, d_2\}$ is claimed to be false, things are different. Here it is not sufficient that either $\not\models d_1$ or $\not\models d_2$; one has to be sure that *all* disambiguations are false, i.e., $\not\models d_1$ and $\not\models d_2$. To model this distribution of falsity, van Eijck and Jaspars [EJ96] use the notions of a countermodel and a falsification relation $\dashv$. Roughly, if only unambiguous expressions appear as premises or consequences $\dashv$ corresponds to $\not\models$, but if at least one underspecified expression appears as premise or consequence, we have to define the (counter-) consequence relation appropriately.

**Definition 6.** We define the *underspecified consequence relation* $\models_u$ and *underspecified falsification relation* $\dashv_u$ for $\mathcal{L}^u$ and an arbitrary model $M$.

1. $M \models_u \varphi$ iff $M \models \varphi$, if $\varphi$ is an unambiguous expression.
   $M \dashv_u \varphi$ iff $M \not\models \varphi$, if $\varphi$ is an unambiguous expression.
2. $M \models_u UR$ iff $M \models d$, for all $d \in \delta(UR)$.
   $M \dashv_u UR$ iff $M \not\models d$, for all $d \in \delta(UR)$.
3. $M \models_u \neg\varphi$ iff $M \dashv_u \varphi$
   $M \dashv_u \neg\varphi$ iff $M \models_u \varphi$
4. $M \models_u \varphi \wedge \psi$ iff $M \models_u \varphi$ and $M \models_u \psi$
   $M \dashv_u \varphi \wedge \psi$ iff $M \dashv_u \varphi$ or $M \dashv_u \psi$
5. $M \models_u \varphi \vee \psi$ iff $M \models_u \varphi$ or $M \models_u \psi$
   $M \dashv_u \varphi \vee \psi$ iff $M \dashv_u \varphi$ and $M \dashv_u \psi$

6. $M \models_u \varphi \to \psi$ iff $M =\!|_u \varphi$ or $M \models_u \psi$
   $M =\!|_u \varphi \to \psi$ iff $M \models_u \varphi$ and $M =\!|_u \psi$
7. $M \models_u \forall x\varphi$ iff $M \models_u \varphi[a]$, for all $a \in D(M)$.
   $M =\!|_u \forall x\varphi$ iff $M =\!|_u \varphi[a]$, for some $a \in D(M)$.
8. $M \models_u \exists x\varphi$ iff $M \models_u \varphi[a]$, for some $a \in D(M)$.
   $M =\!|_u \exists x\varphi$ iff $M =\!|_u \varphi[a]$, for all $a \in D(M)$.

*Example 7.* We now give an example demonstrating the convenience of having the falsification relation.

In our setting of ambiguous expressions, some familiar classical tautologies are no longer valid. For instance, if $A$ is ambiguous and $B$ unambiguous we do not want $(A \wedge B) \to A$ because the two occurrences of $A$ may be disambiguated in different ways. For instance, if $\delta(A) = \{d_1, d_2\}$, then $\models_u (A \wedge B) \to A$ iff $\models_u (d_1 \wedge B) \to d_1, \models_u (d_1 \wedge B) \to d_2, \models_u (d_2 \wedge B) \to d_1$ and $\models_u (d_2 \wedge B) \to d_2$. If we were to model falsity by $\not\models$, applying the definitions would yield:

$$\models_u (A \wedge B) \to A \text{ iff } \not\models_u A \wedge B \text{ or } \models_u A$$
$$\text{iff } \not\models_u A \text{ or } \not\models B \text{ or } \models_u A$$
$$\text{iff } \not\models d_1 \text{ or } \not\models d_2 \text{ or } \not\models B \text{ or } (\models d_1 \text{ and } \models d_2).$$

The latter is classically valid, and it would therefore make the classical tautology valid. On the other hand, if we model falsity by $=\!|_u$ we manage to avoid this, as $=\!|_u$ distributes over disambiguations of $A$, whereas $\not\models$ does not:

$$\models_u (A \wedge B) \to A \text{ iff } =\!|_u A \wedge B \text{ or } \models_u A$$
$$\text{iff } =\!|_u A \text{ or } =\!| B \text{ or } \models_u A$$
$$\text{iff } (\not\models d_1 \text{ and } \not\models d_2) \text{ or } =\!| B \text{ or } (\models d_1 \text{ and } \models d_2).$$

**Definition 8.** Let $\varphi_1, \ldots, \varphi_n, \psi$ be $\mathcal{L}^u$-formulas, possibly containing underspecified representations. We define relation of *underspecified consequence* $\models_u$ as follows:

$$\varphi_1, \ldots, \varphi_n \models_u \psi \text{ iff}$$
$$\text{for all } d_1 \in \delta(\varphi_1), \ldots, d_n \in \delta(\varphi_n)$$
$$\text{and for all } d' \in \delta(\psi) \text{ it holds that}$$
$$d_1, \ldots, d_n \models d'.$$

The underlying intuition is that if someone utters a statement of the form *if S then S'*, where $S$ and $S'$ are ambiguous sentences with $\delta(UR(S)) = \{d_1, d_2\}$, $\delta(UR(S')) = \{d'_1, d'_2\}$, then we do not know exactly what the speaker had in mind by uttering this. So to be sure that this was a valid utterance, one has to check whether it is valid for every possible combination of disambiguations, i.e., whether each of $d_1 \models d'_1$, $d_1 \models d'_2$, $d_2 \models d'_1$, and $d_2 \models d'_2$ is a valid classical consequence.

Unfortunately, this definition of entailment is not a conservative extension of classical logic. Even the reflexivity principle $A \models A$ fails. For instance, if we take $\delta(UR(S)) = \{d_1, d_2\}$, then $UR(S) \models_u UR(S)$ iff $d_1 \models d_1, d_1 \models d_2, d_2 \models d_1$, and $d_2 \models d_2$, i.e. iff $\models d_1 \leftrightarrow d_2$. As we will show below, this has some clear consequences for our calculus, especially the closure conditions. We refer the reader to [Dee96,Jas97] for alternative definitions of the ambiguous entailment relation.

## 4 An Underspecified Tableaux Calculus

The differentiation between consequence and falsification can be nicely modeled in a labeled tableaux calculus, where the nodes in the tableaux tree are of the form $T : \varphi$ or $F : \varphi$, meaning that we want to construct a model or countermodel for $\varphi$, respectively. Tableaux calculi are especially well suited, because the notion of a countermodel is implicit in the notion of an open tableaux tree, where one constructs a countermodel for a formula.

But what does it mean, if we not only allow first-order formulas to appear in a tableaux proof but as also u-formulas? According to the semantic definitions in Section 3, a proof for a u-formula is simply a proof for each of its disambiguations (in a classical tableaux calculus $\mathcal{TC}$). In the following two subsections we first introduce a calculus $\mathcal{TC}_u$ which integrates the mechanism of disambiguation in its deduction rules, and thereby allows one to postpone the disambiguation until it is really needed. $\mathcal{TC}_u$ nicely shows how ambiguity and branching of tableaux trees correspond to each other. But $\mathcal{TC}_u$ still makes no use of the compact representation of underspecified representations, introduced in Section 2. Therefore, we give a modified version of $\mathcal{TC}_u$, called $\mathcal{TC}_{up}$, which also allows us to reason within an underspecified representation.

Our tableaux calculi are based on the labeled free-variable tableaux calculus, see for instance [Fit96] for a general introduction to tableaux calculi.

### 4.1 Reasoning with Total Disambiguations

The definitions of the logical connectives in section 3 allow us to treat logical connectives occurring in u-formulas in the same way as in a tableaux calculus for classical logic $\mathcal{TC}$, as long as they do not occur inside of a *UR*. Here it is necessary to disambiguate the *UR* first, and then apply the rules in the normal way.

*Example 9.* If we try to deduce $(A \wedge B) \to A$, with $\delta(A) = \{d_1, d_2\}$ and $B$ unambiguous, we have to prove each of $\vdash_{\mathcal{TC}} (d_1 \wedge B) \to d_1, \vdash_{\mathcal{TC}} (d_1 \wedge B) \to d_2, \vdash_{\mathcal{TC}} (d_2 \wedge B) \to d_1$ and $\vdash_{\mathcal{TC}} (d_2 \wedge B) \to d_2$. This leads to the following classical labeled tableaux proof trees.

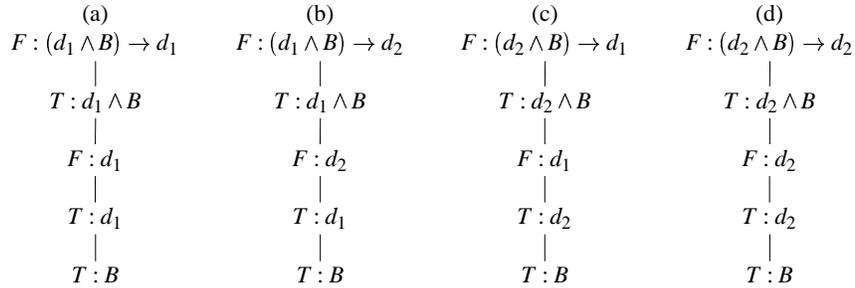

At least structurally, the above proof trees are the same. It does not matter whether they contain underspecified representations. This suggests a natural strategy: to postpone disambiguation and merge those parts of the trees that are similar.

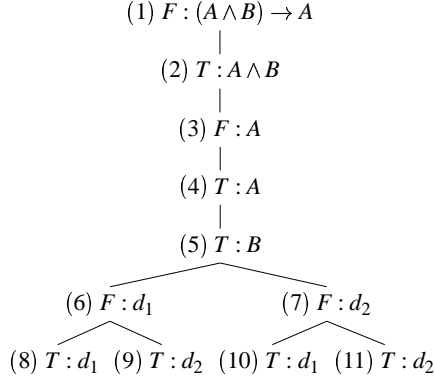

This is a much more compact representation. Again, since *A* is ambiguous, (3) and (4) do not allow one to close the branch, because reflexivity is not a valid principle in our ambiguous setting.

The deduction rules for our underspecified tableaux calculus for totally disambiguated expressions $\mathcal{TC}_u$ are given in Table 1. Besides the last two rules ($T_u$ :*UR*) and ($F_u$ :*UR*), all rules are stated in a standard way and need no further explanation. The purpose of the last two rules is to disambiguate *UR*'s and to start a new branch for each of its disambiguations. This implements the idea of postponing disambiguation, because disambiguation applies now only to *UR*'s and not to any u-formula.

**Theorem 10.** *Let $\varphi \in \mathcal{L}^u$. Then $\vdash_{\mathcal{TC}_u} \varphi$ iff $\vdash_{\mathcal{TC}} d$, for all $d \in \delta(\varphi)$.*

**Corollary 11.** *Let $\varphi \in \mathcal{L}^u$. Then $\vdash_{\mathcal{TC}_u} \varphi$ iff $\models_u \varphi$.*

### 4.2 Reasoning with Partial Disambiguations

From a computational point of view ($T_u : UR$) and ($F_u : UR$) are not optimal, since they cause a lot of branchings of the tableaux tree. Also, total disambiguation is not the appropriate means for underspecified reasoning, because the advantage of the compact representation, namely avoiding redundancy, gets lost. So $\mathcal{TC}_u$ is appropriate for dealing with formulas containing *UR*'s but not for reasoning inside the *UR*'s themselves.

Sometimes it is not necessary to compute all disambiguations, because there exists a strongest (weakest) partial disambiguation. If such a strongest (weakest) disambiguation does exist, it suffices to verify (falsify) this one, because it entails (is entailed by) all other disambiguations. But what are the circumstances under which a strongest (weakest) disambiguation exists?

Before we can determine a strongest (weakest) reading, we have to resolve the relative position of negative contexts and quantifiers. To this end we define positive and negative contexts (see also [TS96]).

**Definition 12.** A u-formula $\varphi$ is a *positive context* for a subformula $\xi$ of $\varphi$, notation: $\text{con}^+(\varphi,\xi)$, iff

$$\varphi ::= \xi \mid \psi \wedge \chi[\xi] \mid \chi[\xi] \wedge \psi \mid \psi \vee \chi[\xi] \mid \chi[\xi] \vee \psi \mid \psi \to \chi[\xi] \mid \forall x \chi[\xi] \mid \exists x \chi[\xi]$$

**Table 1.** Deduction rules of the underspecified tableaux calculus $\mathcal{TC}_u$

$$\frac{T_u : \varphi \land \psi}{\begin{array}{c} T_u : \varphi \\ T_u : \psi \end{array}} \ (T_u : \land) \qquad \frac{F_u : \varphi \land \psi}{F_u : \varphi \ \big| \ F_u : \psi} \ (F_u : \land)$$

$$\frac{T_u : \varphi \lor \psi}{T_u : \varphi \ \big| \ T_u : \psi} \ (T_u : \lor) \qquad \frac{F_u : \varphi \lor \psi}{\begin{array}{c} F_u : \varphi \\ F_u : \psi \end{array}} \ (F_u : \lor)$$

$$\frac{T_u : \varphi \to \psi}{F_u : \varphi \ \big| \ T_u : \psi} \ (T_u : \to) \qquad \frac{F_u : \varphi \to \psi}{\begin{array}{c} T_u : \varphi \\ F_u : \psi \end{array}} \ (F_u : \to)$$

$$\frac{T_u : \neg \varphi}{F_u : \varphi} \ (T_u : \neg) \qquad \frac{F_u : \neg \varphi}{T_u : \varphi} \ (F_u : \neg)$$

$$\frac{T_u : \forall x \varphi}{T_u : \varphi[x/X]} \ (T_u : \forall) \qquad \frac{F_u : \forall x \varphi}{F_u : \varphi[x/f(X_1, \ldots, X_n)]} \ (F_u : \forall)^{\dagger}$$

$$\frac{T_u : \exists x \varphi}{T_u : \varphi[x/f(X_1, \ldots, X_n)]} \ (T_u : \exists)^{\dagger} \qquad \frac{F_u : \exists x \varphi}{F_u : \varphi[x/X]} \ (F_u : \exists)$$

$$\frac{T_u : UR}{T_u : d_1 \ \big| \ \ldots \ \big| \ T_u : d_n} \ (T_u : UR)^{\ddagger} \qquad \frac{F_u : UR}{F_u : d_1 \ \big| \ \ldots \ \big| \ F_u : d_n} \ (F_u : UR)^{\ddagger}$$

$^{\dagger}$Where $X_1, \ldots, X_n$ are the free variables in $\varphi$.
$^{\ddagger}$Where $d_1, \ldots, d_n \in \delta(UR)$.

where $\xi$ occurs in $\chi$ and $con^+(\chi, \xi)$ holds, or $\varphi ::= \neg \chi[\xi] \mid \chi[\xi] \to \psi$, where $\xi$ occurs in $\chi$ and $con^-(\chi, \xi)$ holds.

A u-formula $\varphi$ is a *negative context* for a subformula $\xi$ of $\varphi$, $con^-(\varphi, \xi)$, iff

$$\varphi ::= \psi \land \chi[\xi] \mid \chi[\xi] \land \psi \mid \psi \lor \chi[\xi] \mid \chi[\xi] \lor \psi \mid \psi \to \chi[\xi] \mid \forall x \chi[\xi] \mid \exists x \chi[\xi],$$

where $\xi$ occurs in $\chi$ and $con^-(\chi, \xi)$ holds, or $\varphi ::= \neg \chi[\xi] \mid \chi[\xi] \to \psi$, where $\xi$ occurs in $\chi$ and $con^+(\chi, \xi)$ holds.

To apply the tableaux rules to a formula $\psi$ it is necessary to know whether $\psi$ occurs positively in a superformula $\varphi$ — then we have to apply a *T*-rule —, or negatively — then we have to apply an *F*-rule. In an underspecified representation it may happen that a formula occurs positively in one disambiguation and negatively in another. We call formulas of this kind *indefinite*, and in this case we cannot apply a tableaux rule.

**Definition 13.** Given an underspecified representation $\langle LHF, C, L, H \rangle$, a labeled h-formula $l : \varphi[h] \in LHF$ is *definite* if for every $l' : \psi[h'] \in LHF$, such that $con^-(\psi, h')$ holds and $h \sqcup h'$ defined, then it holds that $l \leq h' \in C$ or $l' \leq h \in C$. It is called *indefinite* otherwise.

Why do we consider definite formulas? Intuitively, we need to know which quantifier we are actually dealing with when we are trying to find a strongest (weakest) reading. Formulas can be made more definite by using the rules for partial negation resolution given in Table 2. Roughly, we obtain more definite h-formulas within a given underspecified representation by adding further constraints which let indefinite h-formulas become definite by using one of the rules of *partial negation resolution* as specified in Table 2, which are generalizations of the method of partial disambiguation in [KR96]. These rules reduce the number of indefinite h-formulas occurring in an underspecified representation by creating partial disambiguations in which the indefinite h-formula has scope over (or is in the scope of one of) the h-formulas inducing the indefiniteness; in Table 2 this is $l_m : \varphi_m[h_n]$, where $\text{con}^-(\varphi_m, h_n)$ holds and $h_k \sqcup h_n$ is defined. Solid lines between two labels or holes, $k$, $k'$, indicate immediate scope relation, dashed lines are the transitive closure of solid lines. For instance, let $\varphi_j = \forall x(\varphi)$ and $\varphi_m = \neg h_n$, we do not know, whether $\forall x$ binds $x$ universally or existentially, because it can appear above or under the negation. Applying $(T_u : \pi)$ yields the two possible cases, namely $\forall x(\varphi)$ occurring above (left branch) or under (right branch) the negation.

To put it differently, suppose that $l_m : \varphi_m[h_n]$ is the only h-formula, which causes indefiniteness of $l_j : \varphi_j$ in an application of $(T_u : \pi)$, then the rule for left partial disambiguation labels $l_j : \varphi_j$ with $T_u$, because now it has scope over the negative context, and the rule for right partial disambiguation labels $l_j : \varphi_j$ with $F_u$, because it is in the scope of the negative context.

Our complete set of deduction rules for underspecified representations is given by combining Tables 2 and 3. This set defines our tableaux calculus, $\mathcal{TC}_{up}$.

Observe that there are three sets of rules in Table 3. The first set deals with ordinary logical connectives only. The second group are so-called interface rules; roughly speaking, they control the flow of information between traditional tableaux reasoning and disambiguation. Reasoning within an underspecified representation starts at its top-hole and compares all its daughters, i.e., those formulas that appear immediately in its scope. A similar interface is needed for h-formulas. The logical connectives in complex h-formulas are also treated with the T/F-rules, but for treating holes we need to know what material goes into them. For holes having only one daughter, it is possible to apply the normal tableaux rules to this daughter, see $(T_u : \uparrow)$ and $(F_u : \uparrow)$.

As to the rules in the third group, these are designed to partially construct the weakest or strongest readings of u-formulas, respectively. Both $(T_u : \forall)$ and $(F_u : \exists)$ presuppose that $l_j : \exists x \varphi[h_l]$ or $l_j : \forall x \varphi[h_l]$ occurs definite, otherwise we would not be able to tell what the quantificational force of $l_j : \exists x \varphi$ or $l_j : \forall x \varphi$ is. So, before applying the rules it may be necessary to apply partial negation resolution as presented in Table 2 first so as to make $l_j : \forall x \varphi[h_l]$ definite. There is an important restriction on the applicability of the rules $(T_u : \forall)$ and $(F_u : \exists)$: to guarantee soundness of the rules, the formulas $\forall x \varphi[h]$ and $\exists x \varphi[h]$ in $l_j$ should be *special*. Here $\forall x \varphi[h]$ is special if it is of the form $\forall x (\chi_1 \rightarrow h)$ or $\forall x (\chi_1 \wedge h \rightarrow \chi_2)$, while $\exists x \varphi[h]$ is special if it is of the form $\exists x (\chi_1 \wedge h)$.

To conclude this section, we briefly turn to soundness and completeness. First, now that our tableaux may have different kinds of labelings (there are $T/F$-nodes and $T_u/F_u$-nodes), we need to specify what it means for a tableaux to close. We say that a branch $b$ *closes* if there are two nodes $T : \varphi$ and $F : \psi$ belonging to $b$, such that $\varphi$ and $\psi$ are atomic formulas of $\mathcal{L}$ and $\varphi$ and $\psi$ are unifiable. In particular, it is not possible to close a tableau with two nodes $T : \varphi$ and $F : \psi$ containing holes or underspecified representations.

**Table 2.** Tableaux rules for partial negation resolution

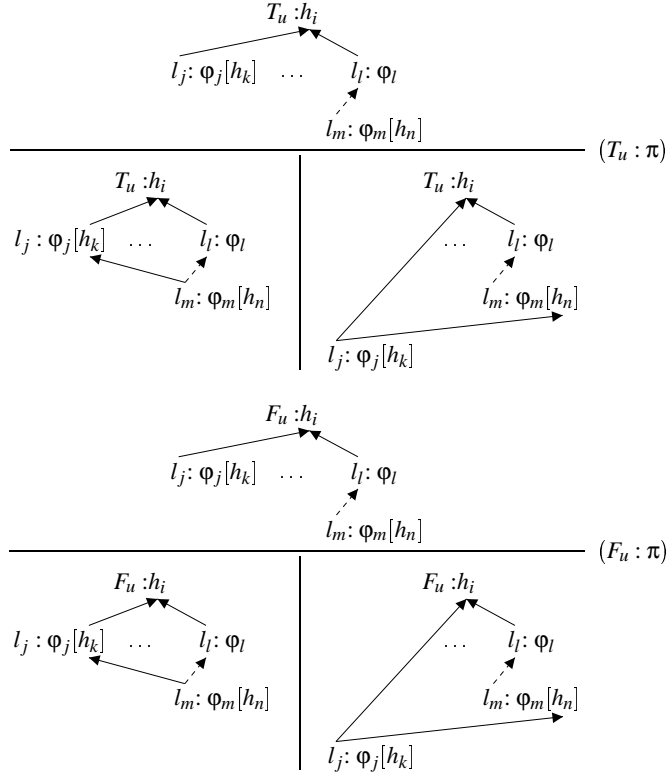

Next, what do soundness and completeness mean in our ambiguous setting? Sound and complete with respect to which semantics or system? We have opted to state soundness and completeness with respect to tableaux provability of all total disambiguations.

**Theorem 14 (Soundness and Completeness).** *Let $\varphi \in \mathcal{L}^u$. Then $\vdash_{\mathcal{TC}_{up}} \varphi$ if, and only if, for all $d \in \delta(\varphi)$ $\vdash_{\mathcal{TC}} d$*

*Proof (Sketch).* The soundness part ('only if') boils down to a proof that the $T_u/F_u$ rules do not introduce any information that would not have been available by totally disambiguating first. The restrictions on the rules $(T_u : \forall)$ and $(F_u : \exists)$ that were discussed above allow us to establish this.

Proving completeness ('if') is in some way easier: any open branch in a (completely developed) tableau for $\mathcal{TC}_{up}$ corresponds to a (completely developed) open branch in a tableau proof for $\mathcal{TC}_u$. See [MR98] for the details.

**Table 3.** Set of deduction and interface rules of $\mathcal{TC}_{up}$

$$\frac{T:\varphi\wedge\psi}{\begin{array}{c}T:\varphi\\T:\psi\end{array}}\;(T:\wedge) \qquad \frac{F:\varphi\wedge\psi}{F:\varphi\;\mid\;F:\psi}\;(F:\wedge)$$

$$\frac{T:\varphi\vee\psi}{T:\varphi\;\mid\;T:\psi}\;(T:\vee) \qquad \frac{F:\varphi\vee\psi}{\begin{array}{c}F:\varphi\\F:\psi\end{array}}\;(F:\vee)$$

$$\frac{T:\varphi\to\psi}{F:\varphi\;\mid\;T:\psi}\;(T:\to) \qquad \frac{F:\varphi\to\psi}{\begin{array}{c}T:\varphi\\F:\psi\end{array}}\;(F:\to)$$

$$\frac{T:\neg\varphi}{F:\varphi}\;(T:\neg) \qquad \frac{F:\neg\varphi}{T:\varphi}\;(F:\neg)$$

$$\frac{T:\forall x\varphi}{T:\varphi[x/X]}\;(T:\forall) \qquad \frac{F:\forall x\varphi}{F:\varphi[x/f(X_1,\ldots,X_n)]}\;(F:\forall)^{\dagger}$$

$$\frac{T:\exists x\varphi}{T:\varphi[x/f(X_1,\ldots,X_n)]}\;(T:\exists)^{\dagger} \qquad \frac{F:\exists x\varphi}{F:\varphi[x/X]}\;(F:\exists)$$

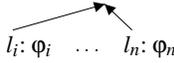

$$\frac{T:UR}{T_u:h_0}\;(T:UR) \qquad \frac{F:UR}{F_u:h_0}\;(F:UR)$$

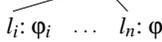

$$\frac{T:h_i}{T_u:h_i}\;(T:h) \qquad \frac{F:h_i}{F_u:h_i}\;(F:h)$$

$$\frac{\begin{array}{c}T_u:h_i\\l_j:\varphi\end{array}}{T:\varphi}\;(T_u:\uparrow) \qquad \frac{\begin{array}{c}F_u:h_i\\l_j:\varphi\end{array}}{F:\varphi}\;(F_u:\uparrow)$$

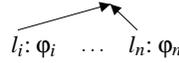

$(T_u:\forall)^{\ddagger}$ $\quad$ $(F_u:\exists)^{\ddagger}$

$^{\dagger}$Where $X_1,\ldots,X_n$ are the free variables in $\varphi$.
$^{\ddagger}$Where $Q\in\{\forall,\exists\}$, $l_j$ is definite, and $\forall x\varphi[h]$ and $\exists x\varphi[h]$ are *special* (see below).

## 5 An Example

Consider the sentence *every boy doesn't see a movie* appearing as a premise in a tableau. Because displaying derivations in our calculus is very space-consuming, we can only give the beginning of one of its branches, which is given in Figure 1. Each box corresponds to a node in a tableau tree. Because in (1) $l_1 : \forall x\, (boy(x) \to h_1)$ occurs indefinite, it is necessary to apply partial negation resolution first. The total disambiguation of the left branching would be

$$\{\forall x\,(boy(x) \to \exists y\,(movie(y) \land \neg see(x,y))),$$
$$\forall x\,(boy(x) \to \neg \exists y\,(movie(y) \land see(x,y))),$$
$$\exists y\,(movie(y) \land \forall x\,(boy(x) \to \neg see(x,y)))\},$$

That is, formulas in which the universal quantifier has scope over the negation, disregarding the existential quantifier. Now $(T_u : \forall)$ is applicable and the universal quantifier is given wide scope in (4), corresponding to the readings $\forall x\,(boy(x) \to \exists y\,(movie(y) \land \neg see(x,y)))$ and $\forall x\,(boy(x) \to \neg \exists y\,(movie(y) \land see(x,y)))$. Because $h_0$ has only one daughter, the normal tableaux rules for logical connectives can be applied to it. So we instantiate $x$ with a free variable $X$ and apply $(T : \to)$, which causes a branching of the proof tree, where (7) is a nonambiguous literal with which we can try to close a tableaux branch. In (8) $h_1$ is the top-node to which the underspecified tableaux rules can be applied.

## 6 Conclusion

In this paper we have presented a tableaux calculus for reasoning with ambiguous quantification. We have set up a representation formalism that allows for a smooth interleaving of traditional deduction steps with disambiguation steps.

Our ongoing work focuses on two aspects. First, we are adding rules for coping with additional forms of ambiguity to the calculus, such as ambiguity of binary connectives. Second, we are in the process of implementing the calculus $\mathcal{TC}_{up}$; as part of this work new and interesting theoretical issues arise, such as 'proof optimization': for reasons of efficiency it pays to postpone disambiguations as long as possible, but to be able to apply some of the rules expressions need to be definite and for this reason early disambiguation may be required. What is the best way of reconciling these two demands?

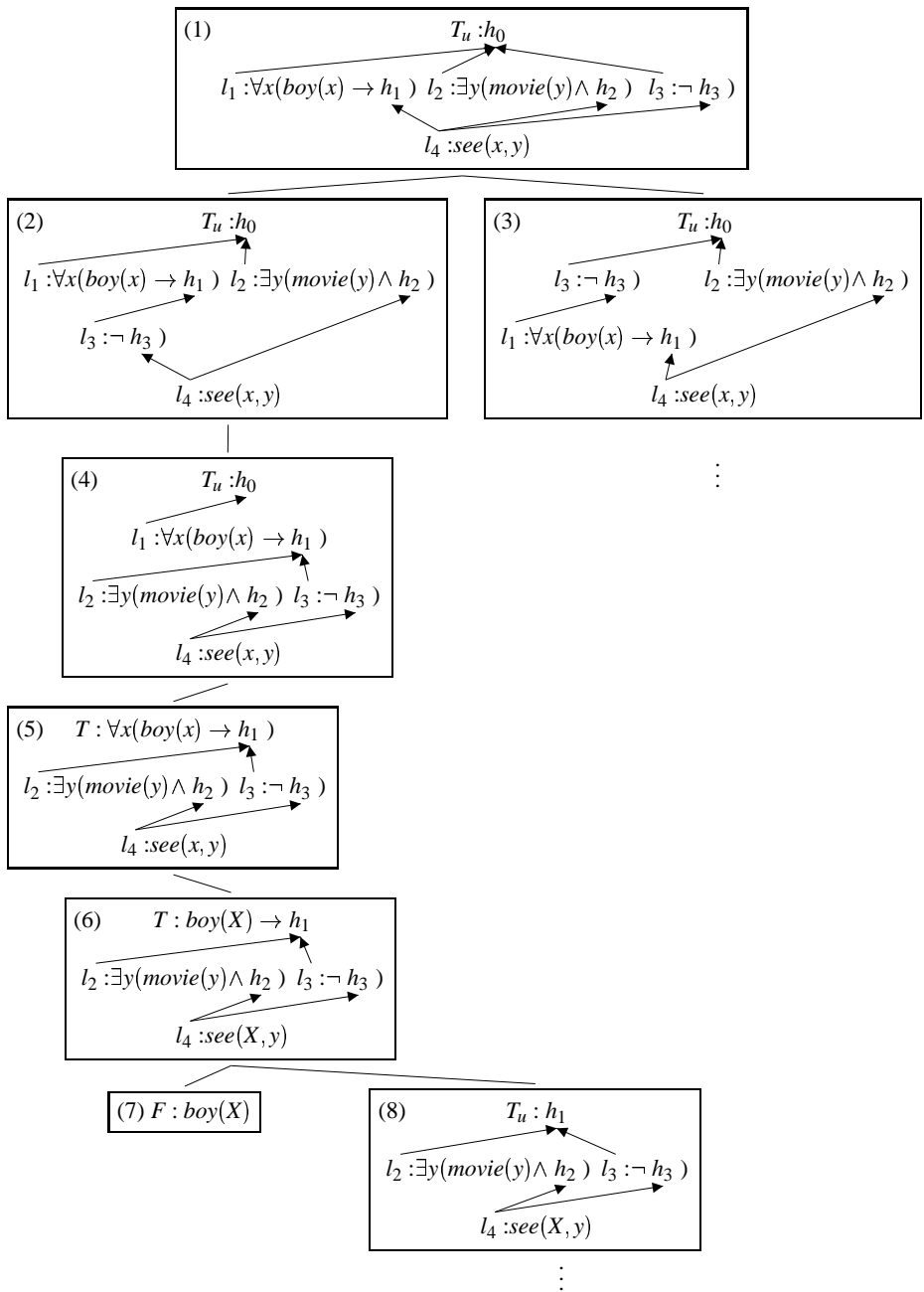

**Fig. 1.** Part of a proof in $\mathcal{TC}_{up}$